\documentclass[letterpaper]{article} 
\usepackage{aaai2026}  
\usepackage{times}  
\usepackage{helvet}  
\usepackage{courier}  
\usepackage[hyphens]{url}  
\usepackage{graphicx} 
\usepackage{natbib}  
\usepackage{caption} 
\usepackage{algorithm}
\usepackage{algorithmic}
\usepackage{amsmath}
\usepackage{multirow}
\usepackage{array}
\usepackage{tabularx}
\usepackage{tcolorbox}
\usepackage{newfloat}
\usepackage{listings}
\DeclareCaptionStyle{ruled}{labelfont=normalfont,labelsep=colon,strut=off} 
\floatstyle{ruled}
\newfloat{listing}{tb}{lst}{}
\floatname{listing}{Listing}
\title{TruthfulRAG: Resolving Factual-level Conflicts in Retrieval-Augmented Generation with Knowledge Graphs}
\author{
    Shuyi Liu, Yuming Shang, Xi Zhang\thanks{Corresponding author.}  
}
\affiliations{
    Key Laboratory of Trustworthy Distributed Computing and Service (MoE) \\

    Beijing University of Posts and Telecommunications, China \\
    \{liushuyi111, shangym, zhangx\}@bupt.edu.cn
}

\usepackage{bibentry}

\begin{document}

\maketitle

\begin{abstract}
Retrieval-Augmented Generation (RAG) has emerged as a powerful framework for enhancing the capabilities of Large Language Models (LLMs) by integrating retrieval-based methods with generative models. As external knowledge repositories continue to expand and the parametric knowledge within models becomes outdated, a critical challenge for RAG systems is resolving conflicts between retrieved external information and LLMs' internal knowledge, which can significantly compromise the accuracy and reliability of generated content. However, existing approaches to conflict resolution typically operate at the token or semantic level, often leading to fragmented and partial understanding of factual discrepancies between LLMs' knowledge and context, particularly in knowledge-intensive tasks. To address this limitation, we propose TruthfulRAG, the first framework that leverages Knowledge Graphs (KGs) to resolve factual-level knowledge conflicts in RAG systems. Specifically, TruthfulRAG constructs KGs by systematically extracting triples from retrieved content, utilizes query-based graph retrieval to identify relevant knowledge, and employs entropy-based filtering mechanisms to precisely locate conflicting elements and mitigate factual inconsistencies, thereby enabling LLMs to generate faithful and accurate responses. Extensive experiments reveal that TruthfulRAG outperforms existing methods, effectively alleviating knowledge conflicts and improving the robustness and trustworthiness of RAG systems.
\end{abstract}


\section{Introduction}

Large Language Models (LLMs) have demonstrated impressive performance across diverse natural language understanding and generation tasks~\cite{achiam2023gpt,touvron2023llama,yang2025qwen3}. Despite their proficiency, LLMs remain ineffective in handling specialized, privacy-sensitive, or time-sensitive knowledge that is not encompassed within their training corpora~\cite{zhang2024knowgpt,huang2025survey}. For the solutions, Retrieval-Augmented Generation (RAG) has emerged as a promising paradigm that enhances the relevance and factuality of the generated responses by integrating external knowledge retrieval with the remarkable generative capabilities of LLMs~\cite{lewis2020retrieval,gao2023retrieval,fan2024survey}. However, as RAG systems continuously update their knowledge repositories, the temporal disparity between dynamic external sources and static parametric knowledge within LLMs inevitably leads to knowledge conflicts~\cite{xie2023adaptive,xu2024knowledge,shi2024ircan}, which can significantly undermine the accuracy and reliability of the generated content.

\begin{figure}[t]
\centering
\includegraphics[width=0.48\textwidth]{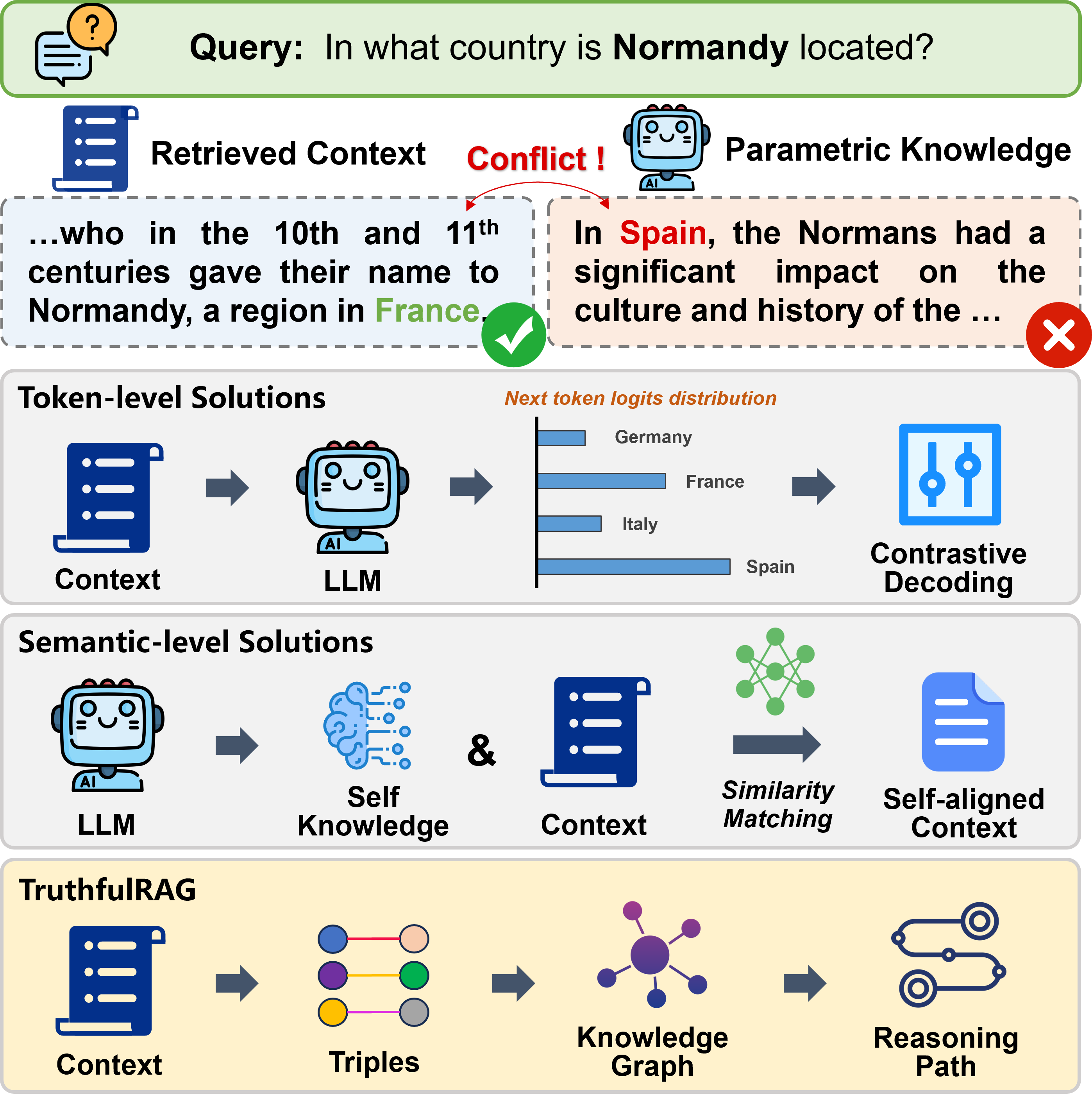} 
\caption{The illustration of knowledge conflicts and the differences between existing solutions and TruthfulRAG.}
\label{fig:diff}
\end{figure}

Recent research has begun to investigate the impact of knowledge conflicts on the performance of RAG systems~\cite{chen2022rich,xie2023adaptive,tan2024blinded} and explore methods to mitigate such conflicts~\cite{wang2024astute,jin2024tug,zhang2025faithfulrag,bi2025parameters}. Existing resolution approaches can be categorized into two methodological types: (i) token-level methods, which manage LLMs' preference between internal and external knowledge by adjusting the probability distribution over the output tokens~\cite{jin2024tug,bi2025parameters}; (ii) semantic-level methods, which resolve conflicts by semantically integrating and aligning knowledge segments from internal and external sources~\cite{wang2024astute,zhang2025faithfulrag}. However, these token-level or semantic-level conflict resolution methods generally employ coarse-grained strategies that rely on fragmented data representations, resulting in insufficient contextual awareness. This may prevent LLMs from accurately capturing complex interdependencies and fine-grained factual inconsistencies, especially in knowledge-intensive conflict scenarios~\cite{han2024retrieval}.

To address the above limitations, we propose TruthfulRAG, the first framework that leverages Knowledge Graphs (KGs) to resolve factual-level conflicts in RAG systems. As illustrated in Figure \ref{fig:diff}, unlike previous studies, TruthfulRAG uses structured triple-based knowledge representations to construct reliable contexts, thereby enhancing the confidence of LLMs in external knowledge and facilitating trustworthy reasoning. The TruthfulRAG framework comprises three key modules: (a) Graph Construction, which derives structured triples from retrieved external knowledge by identifying entities, relations, and attributes to construct knowledge graphs; (b) Graph Retrieval, which conducts query-based retrieval algorithms to obtain relevant knowledge that exhibit strong factual associations with the input query; and (c) Conflict Resolution, which applies entropy-based filtering techniques to locate conflicting elements and mitigate factual inconsistencies, ultimately forming more reliable reasoning paths and promoting more accurate outputs. This framework integrates seamlessly with existing RAG architectures, enabling the extraction of highly relevant and factually consistent knowledge, effectively eliminating factual-level conflicts and improving generation reliability.

The contributions of this paper are as follows:
\begin{itemize}
\item We discover that constructing contexts through textual representations on structured triples can enhance the confidence of LLMs in external knowledge, thereby promoting trustworthy and reliable model reasoning.
\item We introduce TruthfulRAG, the first framework that leverages knowledge graphs to resolve factual-level conflicts in RAG systems through systematic triple extraction, query-based graph retrieval, and entropy-based filtering mechanisms.
\item We conduct extensive experiments demonstrating that TruthfulRAG outperforms existing methods in mitigating knowledge conflicts while improving the robustness and trustworthiness of RAG systems.
\end{itemize}

\begin{figure*}[t]
\centering
\includegraphics[width=1\textwidth]{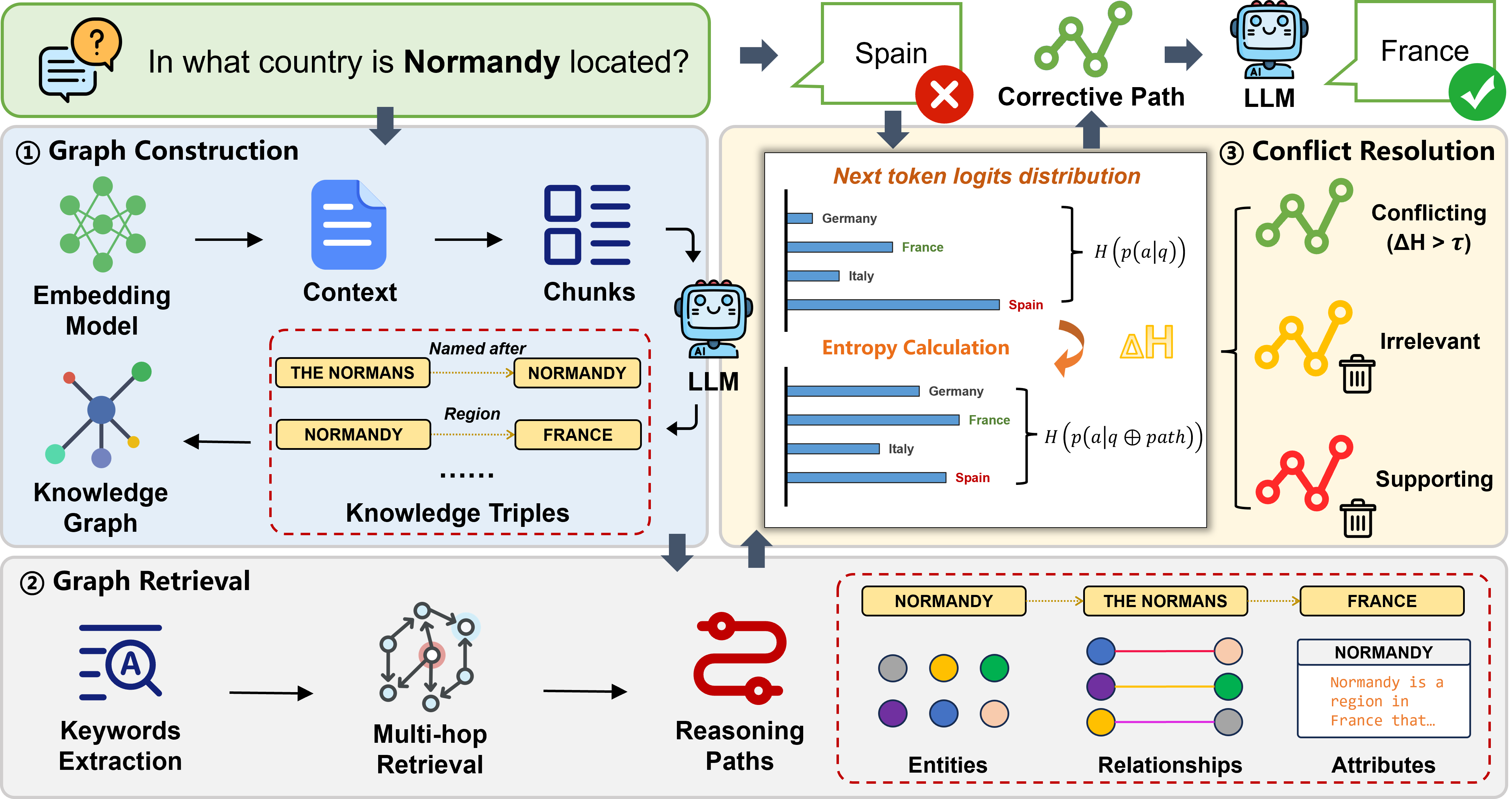} 
\caption{The overall pipeline of the TruthfulRAG framework. TruthfulRAG first extracts structured knowledge triples to construct a comprehensive knowledge graph. Subsequently, it employs query-aware graph traversal to identify salient reasoning paths, where each path comprises entities and relationships enriched with associated attributes. Finally, the framework applies entropy-based conflict resolution to detect and filter out corrective paths that challenge parametric misconceptions, thereby alleviating knowledge conflicts between internal and external information, prompting consistent and credible responses.}
\label{fig:main}
\end{figure*}

\section{Methodology}
In this section, we provide a detailed introduction to the TruthfulRAG framework. As illustrated in Figure~\ref{fig:main}, TruthfulRAG comprises three interconnected modules: (i) Graph Construction, which transforms unstructured retrieved content into structured knowledge graphs through systematic triple extraction; (ii) Graph Retrieval, which employs query-aware graph traversal algorithms to identify semantically relevant reasoning paths; and (iii) Conflict Resolution, which utilizes entropy-based filtering mechanisms to detect and mitigate factual inconsistencies between parametric and external knowledge.

\subsection{Graph Construction}
The construction of a knowledge graph begins with the conversion of raw information retrieved from the RAG system into structured knowledge representations through systematic entity-relation-attribute extraction. 

Given the retrieved content $C$ for the user's query $q$, we first perform fine-grained semantic segmentation to partition the content into coherent textual segments $\mathcal{S}=\{s_1, s_2, \ldots, s_m\}$, where each segment $s_i$ represents a semantically coherent unit containing factual information.
For each textual segment $s_i\in\mathcal{S}$, we employ the generative model $\mathcal{M}$ from the RAG system to extract a set of structured knowledge triples $\mathcal{T}_{all} = \{\mathcal{T}_{i,1}, \mathcal{T}_{i,2}, \ldots, \mathcal{T}_{i,n}\}$, with each triple $\mathcal{T}_{i,j} = (h, r, t)$ consisting of a head entity $h$, relation $r$, tail entity $t$. This extraction process aims to capture both explicit factual statements and implicit semantic relationships embedded within the original content, thereby ensuring the comprehensiveness and semantic integrity of the knowledge representation.

The aggregated triple set from all retrieved content forms the foundation for constructing the knowledge graph $\mathcal{G}$:
\begin{equation}
\mathcal{G} = (\mathcal{E}, \mathcal{R}, \mathcal{T}_{all})
\end{equation}
where $\mathcal{E} = \bigcup_{i,j,k} {h_{i,j,k}, t_{i,j,k}}$ represents the entity set, $\mathcal{R} = \bigcup_{i,j,k} {r_{i,j,k}}$ denotes the relation set, and $\mathcal{T}_{all} = \bigcup_{i,j} \mathcal{T}_{i,j}$ constitutes the complete triple repository. This structured knowledge representation enables the filtering of low-information noise and captures detailed factual associations, thereby providing a clear and semantically enriched foundation for subsequent query-aware knowledge retrieval.

\subsection{Graph Retrieval}
To acquire knowledge that is strongly aligned with user queries at the factual level, we design a query-aware graph traversal algorithm that can identify critical knowledge paths within the graph, ensuring both semantic relevance and factual consistency in the retrieval process.

Initially, key elements are extracted from the user query $q$ to serve as important references for matching components in the knowledge graph. These elements include the query’s target entities, relations, and intent categories, denoted as $\mathcal{K}_q$. Subsequently, semantic similarity matching is employed to identify the top-$k$ most relevant entities and relations within the knowledge graph:
\begin{align}
\mathcal{E}{imp} &= \text{TopK}({\text{sim}(e, \mathcal{K}_q) : e \in \mathcal{E}}, k) \\
\mathcal{R}{imp} &= \text{TopK}({\text{sim}(r, \mathcal{K}_q) : r \in \mathcal{R}}, k)
\end{align}
where $\text{sim}(\cdot, \cdot)$ represents the semantic similarity function computed using dense embeddings, $\mathcal{E}{imp}$ denotes the set of key entities, and $\mathcal{R}{imp}$ represents the set of key relations.
From each key entity $e \in \mathcal{E}{imp}$, we perform a two-hop graph traversal to systematically collect the entire set of possible initial reasoning paths $\mathcal{P}{init}$. 

To further filter reasoning paths with stronger factual associations, we introduce a fact-aware scoring mechanism that evaluates the relevance of paths to the query based on the coverage of key entities and relations within each path p:
\begin{equation}
\text{Ref}(p) = \alpha \cdot \frac{|{e \in p} \cap \mathcal{E}{imp}|}{|\mathcal{E}{imp}|} + \beta \cdot \frac{|{r \in p} \cap \mathcal{R}{imp}|}{|\mathcal{R}{imp}|}
\end{equation}
where $\alpha$ and $\beta$ are hyperparameters that control the relative importance of entity and relationship coverage, respectively. The top-scored reasoning paths from $\mathcal{P}{init}$ constitute the core knowledge paths $\mathcal{P}{super}$.
\begin{equation}
\mathcal{P}{super} = \text{TopK}({\text{Ref}(p) : p \in \mathcal{P}{init}}, K)
\end{equation}
In order to construct detailed contextual information, each core reasoning path $p \in \mathcal{P}{super}$ will be represented as a comprehensive contextual structure consisting of three essential components:
\begin{equation}
p = \mathcal{C}_{path} \oplus \mathcal{C}_{entities} \oplus \mathcal{C}_{relations}
\end{equation}
where:
\begin{itemize}
\item $\mathcal{C}{path}$ represents the complete sequential reasoning path: $e_1 \xrightarrow{r_1} e_2 \xrightarrow{r_2} \cdots \xrightarrow{r_{n-1}} e_n$, capturing the logical progression of entities connected through relational links.
\item $\mathcal{C}{entities} = {(e, \mathcal{A}{e}) : e \in p \cap \mathcal{E}{imp}}$ encompasses all important entities within the path along with their corresponding attribute descriptions $\mathcal{A}{e}$, providing thorough entity-specific information for the context.
\item $\mathcal{C}{relations} = {(r, \mathcal{A}{r}) : r \in p \cap \mathcal{R}{imp}}$ includes all important relations on the path together with their corresponding attributes $\mathcal{A}{r}$, enriching the semantic and contextual understanding of the relations.
\end{itemize}
This formalized representation of knowledge ensures that each extracted reasoning path preserves structural coherence through the entity-relation sequence and reinforces semantic richness via comprehensive attribute information, thereby facilitating more nuanced and context-aware knowledge integration for subsequent conflict resolution processes.

\subsection{Conflict Resolution}
To address factual inconsistencies between parametric knowledge and external information, ensuring that LLMs consistently follow the retrieved knowledge paths to achieve accurate reasoning, we employ entropy-based model confidence analysis to investigate the influence of conflicting knowledge on model prediction uncertainty, thereby systematically identifying and resolving factual conflicts based on uncertainty quantification mechanisms.

We implement conflict detection by comparing model performance under two distinct conditions: (1) pure parametric generation without access to external context, and (2) retrieval-augmented generation that incorporates structured reasoning paths constructed from knowledge graph.
For parametric-based generation, we calculate the response probability from LLMs as baselines:
\begin{equation}
P_{param}(ans|q) = \mathcal{M}(q)
\end{equation}
where $ans$ represents the generated answer and $\mathcal{M}(q)$ denotes the response distribution of the LLM based solely on query $q$.
For retrieval-augmented generation, we incorporate each reasoning path from $\mathcal{P}{super}$ as contextual information to obtain the model's output probability:
\begin{equation}
P_{aug}(ans|q, p) = \mathcal{M}(q \oplus p), \quad \forall p \in \mathcal{P}{super}
\end{equation}
where $\mathcal{M}(q \oplus p)$ represents the response distribution of the LLM conditioned on the query $q$ and its corresponding reasoning paths extracted from the knowledge graph.

Inspired by previous research on probability-based uncertainty estimation~\cite{arora2021types,duan2024shifting}, we adopt entropy-based metrics to quantify the model's confidence in the retrieved knowledge:
\begin{equation}
H(P(ans|context)) = -\frac{1}{|l|}\sum_{t=1}^{|l|} \sum_{i=1}^{k} pr_i^{(t)} \log_2 pr_i^{(t)}
\end{equation}
where $pr_i^{(t)}$ represents the probability distribution over the top-$k$ candidate tokens at position $t$, and $|l|$ denotes the token length of the answer. Accordingly, we obtain $H(P_{param}(ans|q))$ for parametric generation and $H(P_{aug}(ans|q, p))$ for retrieval-augmented generation incorporating with individual reasoning path $p$. Consequently, we can utilize the entropy variation under different reasoning paths as a characteristic indicator of knowledge conflict:
\begin{equation}
\Delta H_p = H(P_{aug}(ans|q, p)) - H(P_{param}(ans|q))
\end{equation}
where positive values of $\Delta H_p$ indicate that the retrieved external knowledge intensifies uncertainty in the LLM's reasoning, potentially indicating factual inconsistencies with its parametric knowledge, whereas negative values suggest that the retrieved knowledge aligns with the LLM’s internal understanding, thereby reducing uncertainty. Reasoning paths exhibiting entropy changes exceeding a predefined threshold $\tau$ are classified as $\mathcal{P}_{corrective}$:
\begin{equation}
\mathcal{P}{corrective} = {p \in \mathcal{P}{super} : \Delta H_p > \tau}
\end{equation}
These identified corrective knowledge paths, which effectively challenge and potentially rectify the LLM’s internal misconceptions, are subsequently aggregated to construct the refined contextual input. The final response is then generated by the LLM based on the enriched context:
\begin{equation}
\text{Response} = \mathcal{M}(q \oplus \mathcal{P}{corrective})
\end{equation}
This entropy-based conflict resolution mechanism ensures that LLMs consistently prioritize factually accurate external information when generating responses, improving reasoning accuracy and trustworthiness, thereby enhancing the overall robustness of the RAG system.

\begin{table*}[ht]
\centering
\renewcommand\arraystretch{1.1}
\setlength{\tabcolsep}{2mm}
\begin{tabular}{c|c|cccc|>{\centering\arraybackslash}p{1cm}>{\centering\arraybackslash}p{1cm}}
\hline
\multirow{2}{*}{\textbf{Method}} & \multirow{2}{*}{\textbf{LLM}} & \multicolumn{4}{c|}{\textbf{Dataset}} & \multirow{2}{*}{\textbf{Avg.}} & \multirow{2}{*}{\textbf{Imp.}} \\
\cline{3-6}
 & & FaithEval & MuSiQue & RealtimeQA & SQuAD & & \\
\hline
\multirow{3}{*}{w/o RAG} 
    & GPT-4o-mini         & 4.6  & 15.1 & 43.4 & 11.2 & 18.6 & - \\
    & Qwen2.5-7B-Instruct & 4.2  & 19.6 & 40.7 & 11.1 & 18.9 & - \\
    & Mistral-7B-Instruct & 6.3  & 13.8 & 29.2 & 11.5 & 15.2 & - \\
\hline
\multirow{3}{*}{w/ RAG} 
    & GPT-4o-mini         & 61.3 & 72.6 & 67.3 & 73.1 & 68.6 & 50.0 \\
    & Qwen2.5-7B-Instruct & 53.1 & 75.2 & 78.7 & 68.3 & 68.8 & 49.9 \\
    & Mistral-7B-Instruct & 61.9 & 67.6 & 52.2 & 67.2 & 62.2 & 47.0 \\
\hline
\multirow{3}{*}{KRE} 
    & GPT-4o-mini         & 50.7 & 34.6 & 47.5 & 65.3 & 49.5 & 30.9 \\
    & Qwen2.5-7B-Instruct & 59.6 & 70.7 & \textbf{86.7} & 73.7 & 72.7 & 53.8 \\
    & Mistral-7B-Instruct & 73.2 & 50.6 & 76.9 & 74.6 & 68.8 & 53.6 \\
\hline
\multirow{3}{*}{COIECD} 
    & GPT-4o-mini         & 53.9 & 56.4 & 48.7 & 57.6 & 54.2 & 35.6 \\
    & Qwen2.5-7B-Instruct & 62.3 & 69.7 & 78.8 & 70.8 & 70.4 & 51.5 \\
    & Mistral-7B-Instruct & 62.8 & 66.8 & 58.4 & 65.4 & 63.3 & 48.1 \\
\hline
\multirow{3}{*}{FaithfulRAG} 
    & GPT-4o-mini         & \underline{67.2} & \underline{79.3} & \underline{78.8} & \underline{80.8} & \underline{76.5} & \underline{58.0} \\
    & Qwen2.5-7B-Instruct & \underline{71.8} & \underline{78.0} & \underline{84.1} & \underline{78.3} & \underline{78.1} & \underline{59.1} \\
    & Mistral-7B-Instruct & \underline{81.7} & \underline{78.5} & \underline{77.0} & \textbf{85.7} & \underline{80.7} & \underline{65.5} \\
\hline
\multirow{3}{*}{TruthfulRAG (Ours)} 
    & GPT-4o-mini         & \textbf{69.5} & \textbf{79.4} & \textbf{85.0} & \textbf{81.1} & \textbf{78.8} & \textbf{60.2} \\
    & Qwen2.5-7B-Instruct & \textbf{73.2} & \textbf{79.1} & 82.3 & \textbf{78.7} & \textbf{78.3} & \textbf{59.4} \\
    & Mistral-7B-Instruct & \textbf{81.9} & \textbf{79.3} & \textbf{81.4} & \underline{82.7} & \textbf{81.3} & \textbf{66.1} \\
\hline
\end{tabular}
\caption{Comparison of ACC between TruthfulRAG and five baselines across four datasets within three representive LLMs. The best result for each backbone LLM within each dataset is highlighted in \textbf{bold}, and the second best is emphasized with an \underline{underline}. \textbf{Avg.} denotes the arithmetic mean accuracy across the four datasets, while \textbf{Imp.} indicates the average improvement over the corresponding LLM's w/o RAG baseline.}
\label{tab:main}
\end{table*}

\section{Experiments}
In this section, we present comprehensive experiments to evaluate the effectiveness of TruthfulRAG in resolving knowledge conflicts and enhancing the reliability of RAG systems. Specifically, we aim to address the following research questions: (1) How does TruthfulRAG perform compared to other methods in terms of factual accuracy? (2) What is the performance of TruthfulRAG in non-conflicting contexts? (3) To what extent do structured reasoning paths affect the confidence of LLMs compared to raw natural language context? (4) What are the individual contributions of each module within the TruthfulRAG framework?

\subsection{Experimental Setup}

\subsubsection{Datasets}
We conduct experiments on four datasets that encompass various knowledge-intensive tasks and conflict scenarios. 
FaithEval~\cite{ming2025faitheval} is designed to assess whether LLMs remain faithful to unanswerable, inconsistent, or counterfactual contexts involving complex logical-level conflicts beyond the entity level.
MuSiQue~\cite{trivedi2022musique} and SQuAD~\cite{rajpurkar2016squad} come from previous research KRE~\cite{ying2024intuitive}, which contain fact-level knowledge conflicts that necessitate compositional multi-hop reasoning, making it particularly suitable for evaluating knowledge integration and conflict resolution in complex reasoning scenarios.
RealtimeQA~\cite{kasai2023realtime} focuses on temporal conflicts, where answers may quickly become outdated, leading to inconsistencies between static parametric knowledge and dynamic external sources.

\subsubsection{Evaluated Models}
We select three representative LLMs across different architectures and model scales to ensure comprehensive evaluations: GPT-4o-mini~\cite{achiam2023gpt}, Qwen2.5-7B-Instruct~\cite{yang2025qwen3}, and Mistral-7B-Instruct~\cite{jiang2024mixtral}. This selection encompasses both open-source and closed-source models, ensuring that TruthfulRAG is broadly applicable to RAG systems built upon diverse LLM backbones.

\subsubsection{Baselines}
We compare TruthfulRAG against five baseline approaches spanning different methodological categories:
(i) Direct Generation requires LLMs to generate responses solely based on their parametric knowledge without any external retrieval.
(ii) Standard RAG represents the conventional retrieval-augmented generation paradigm, where LLMs generate responses using retrieved textual passages directly.
(iii) KRE~\cite{ying2024intuitive} serves as a representative prompt optimization method, which enhances reasoning faithfulness by adopting specialized prompting strategies to guide the model in resolving knowledge conflicts.
(iv) COIECD~\cite{yuan2024discerning} represents the decoding manipulation category, which modifies the model's decoding strategy during the inference stage to guide LLMs toward greater reliance on retrieved context rather than parametric knowledge.
(v) FaithfulRAG~\cite{zhang2025faithfulrag} incorporates a self-reflection mechanism that identifies factual discrepancies between parametric knowledge and retrieved context, enabling LLMs to reason and integrate conflicting facts before generating content.

\subsubsection{Evaluation Metrics}
Following prior studies, we adopt accuracy (ACC) as the primary evaluation metric, measuring the proportion of questions for which the LLM generates correct answers, thereby providing a direct assessment of the factual correctness of the generated responses. To evaluate the method's capability to precisely extract information pertinent to the target answer from retrieved corpora, we introduce the Context Precision Ratio (CPR) metric, which measures the proportion of answer-related content within the processed context:
\begin{equation}
\text{CPR} = \frac{|\mathcal{A}_{gold} \cap \mathcal{C}_{processed}|}{|\mathcal{C}_{processed}|}
\end{equation}
where $|\text{Context}_{gold}|$ denotes the length of segments directly related to the correct answer, and $|\text{Context}_{processed}|$ represents the total length of the processed context.

\subsubsection{Implementation Details}
For dense retrieval, cosine similarity is computed using embeddings generated by the all-MiniLM-L6-v2. For entropy-based filtering, we set model-specific thresholds $\tau$ for entropy variation $\Delta H_p$: GPT-4o-mini and Mistral-7B-Instruct use $\tau = 1$, while Qwen2.5-7B-Instruct adopts a higher threshold of $\tau = 3$.
All experiments are conducted using NVIDIA V100 GPUs with 32GB memory. To ensure reproducibility, the temperature for text generation is set to 0, and all Top-$ K $ values are set to 10.

\subsection{Results and Analysis}

\subsubsection{Overall Performance}

Table~\ref{tab:main} presents a comprehensive comparison of TruthfulRAG against five baseline methods across four datasets, evaluating performance in terms of factual accuracy (ACC) using three representative LLMs. To facilitate overall assessment, we additionally report \textbf{Avg.}, the arithmetic mean accuracy across the four datasets, and \textbf{Imp.}, the average improvement over the corresponding LLM's w/o RAG baseline, serving as a proxy for the number of factual conflicts successfully corrected by the method from the LLM’s parametric knowledge.

The results clearly demonstrate that TruthfulRAG consistently achieves superior or competitive performance relative to all baseline approaches. Specifically, it achieves the highest accuracy on FaithEval (81.9\%), MuSiQue (79.4\%), and RealtimeQA (85.0\%), and ranks first or second on SQuAD across all models. Notably, TruthfulRAG achieves the highest overall performance across all backbone LLMs, attaining both the best average accuracy (\textbf{Avg.}) and the greatest relative improvement (\textbf{Imp.}) compared to all baseline methods. This clearly illustrates its robustness in mitigating factual inconsistencies that standard RAG systems struggle with due to unresolved evidence conflicts.

Compared to standard RAG systems, which exhibit significant variability in accuracy due to unresolved knowledge conflicts, TruthfulRAG achieves improvements ranging from 3.6\% to 29.2\%, highlighting its robustness in mitigating factual inconsistencies. Furthermore, while methods like FaithfulRAG and KRE offer partial gains through semantic alignment or prompt-based mechanisms, they fall short in consistently resolving fine-grained factual discrepancies. In contrast, TruthfulRAG integrates knowledge graph-based reasoning with entropy-guided conflict filtering mechanisms to identify and resolve contradictory information, thereby substantially enhancing factual reliability. These findings validate the effectiveness of TruthfulRAG in delivering accurate, faithful, and contextually grounded responses across diverse knowledge-intensive tasks.

\begin{table*}[ht]
\centering
\renewcommand\arraystretch{1.4}
\setlength{\tabcolsep}{2mm}
\begin{tabular}{c|cccccc}
\hline
\multirow{2}{*}{\textbf{Dataset}} & \multicolumn{6}{c}{\textbf{Method}} \\
\cline{2-7}
 & w/o RAG & w/ RAG & KRE & COIECD & FaithfulRAG & TruthfulRAG (Ours) \\ 
\hline
MuSiQue-golden & 45.6 & 89.9 & 44.1(-45.8) & 89.5(-0.4) &91.8(+1.9) & \textbf{93.2} \textbf{(+3.3)} \\ \hline
SQuAD-golden & 68.7 & 97.9 & 83.2(-14.7) & 97.1(-0.8) & 98.1(+0.2) & \textbf{98.3} \textbf{(+0.4)} \\ \hline
\end{tabular}
\caption{Performance comparison on non-conflicting contexts with GPT-4o-mini as the backbone LLM. The best result on each dataset is highlighted in \textbf{bold}. The numbers in parentheses indicates the change in accuracy compared to the standard RAG.}
\label{tab:gold}
\end{table*}

\begin{figure*}[ht]
\centering
\includegraphics[width=1\textwidth]{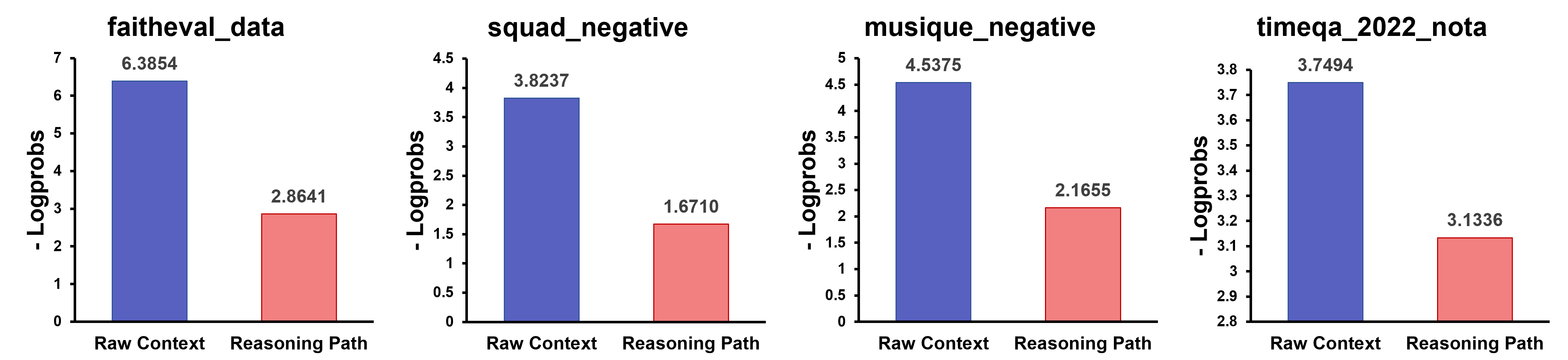} 
\caption{Comparison of LLM confidence, measured by negative log-probability (logprob) values using GPT-4o-mini, when reasoning with natural language contexts versus structured reasoning paths across four datasets. Lower negative logprob values indicate higher actual log-probability scores and thus increased model confidence in generating correct answers.}
\label{fig:confidence}
\end{figure*}

\subsubsection{Performance on Non-Conflicting Contexts}
To evaluate the robustness of TruthfulRAG in scenarios where retrieved contexts free from factual conflicts, we conduct experiments on golden standard datasets in which the retrieved passages are guaranteed to be non-contradictory.

\begin{table*}[ht]
\centering
\renewcommand\arraystretch{1.4}
\setlength{\tabcolsep}{4mm}
\begin{tabular}{c|cccc}
\hline
\multirow{2}{*}{\textbf{Method}} & \multicolumn{4}{c}{\textbf{Dataset}} \\
\cline{2-5}
 & FaithEval & MuSiQue & RealtimeQA & SQuAD \\
\hline
Standard RAG & 61.3 / 0.51 & 72.6 / 1.86 & 67.3 / 0.47 & 73.1 / 2.71 \\
\hline
w/o Knowledge Graph & 64.8 / 0.52 & 78.9 / 1.15 & 83.2 / 0.23 & 78.8 / 1.97 \\
\hline
w/o Conflict Resolution & 69.3 / 0.59 & 77.8 / 2.79 & 84.1 / 1.80 & 78.2 / 2.85 \\
\hline
Full Method & 69.5 / 0.56 & 79.4 / 2.25 & 85.0 / 1.54 & 81.1 / 2.56 \\
\hline
\end{tabular}
\caption{Ablation study results of different components in TruthfulRAG with GPT-4o-mini as the backbone LLM. The results are presented in the format ACC / CPR, where ACC denotes accuracy and CPR represents Context Precision Ratio.}
\label{tab:ablation}
\end{table*}

As shown in Table~\ref{tab:gold}, TruthfulRAG consistently outperforms all baseline methods across both the MuSiQue-golden and SQuAD-golden datasets. These findings substantiate that TruthfulRAG not only excels at resolving conflicting information but also maintains superior performance in non-conflicting contexts, thereby revealing its universal applicability and effectiveness. The consistent performance improvements can be attributed to the structured knowledge representation provided by the knowledge graph module, which enables the identification of fine-grained entities and relational links in non-conflicting contexts. This capability facilitates the extraction of query-relevant information and promotes a more comprehensive understanding and integration of factual knowledge by the LLMs. Notably, while methods such as KRE exhibit significant performance degradation in non-conflicting scenarios, TruthfulRAG maintains its robustness across diverse contextual settings. This consistency highlights its practical utility and reliability for real-world RAG applications.

\subsubsection{Impact of Structured Reasoning Paths}
To investigate the impact of structured reasoning paths on the confidence of LLMs relative to raw natural language context, we conduct a comprehensive analysis across four datasets. Specifically, we compare the model's confidence when reasoning with retrieved knowledge presented in natural language format or as structured reasoning paths derived through our knowledge graph construction mechanism. To quantify the model's confidence in its predicted answers, we measure the log-probability of the correct answer tokens generated by LLMs and compute the average across all test instances. 

As shown in Figure~\ref{fig:confidence}, our experimental results reveal a consistent pattern across all evaluated datasets. Structured reasoning paths consistently lead to higher logprob values for correct answers compared to natural language contexts, indicating greater model confidence when reasoning with structured knowledge representations. This empirical evidence demonstrates that transforming unstructured natural language into structured reasoning paths through knowledge graphs significantly strengthens the LLM's confidence in following external retrieved knowledge for inference. Furthermore, this finding provides crucial insights into the superior performance of TruthfulRAG in both conflicting and non-conflicting semantic scenarios, as the enhanced confidence facilitates more reliable adherence to external knowledge sources, thereby supporting factual consistency and promoting the generation of faithful model outputs.

\subsubsection{Ablation Study}

To comprehensively evaluate the contribution of each component in TruthfulRAG, we conduct systematic ablation experiments by removing key modules from the full framework. Since knowledge graph construction and retrieval are two closely coupled modules, we combine them as an integrated component for ablation evaluation.

As shown in table~\ref{tab:ablation}, the complete TruthfulRAG framework achieves superior performance across all datasets, with accuracy improvements ranging from 6.8\% to 17.7\% compared to the standard RAG, demonstrating that the structured knowledge graph and the conflict resolution mechanism function synergistically to enhance both factual accuracy and contextual precision. The ablation results reveal several critical insights. First, when employing only the filtering mechanism without knowledge graph integration (w/o Knowledge Graph), although accuracy demonstrates modest improvements, CPR exhibits a notable decline across most datasets, particularly in MuSiQue (1.86 to 1.15) and SQuAD (2.71 to 1.97). This phenomenon indicates that LLMs encounter substantial difficulties in effectively extracting relevant information from naturally organized contexts, thereby constraining their ability to achieve higher accuracy. In contrast, when utilizing solely the knowledge graph component without conflict resolution (w/o Conflict Resolution), CPR achieves significant improvements, yet the introduction of extensive structured knowledge simultaneously introduces redundant information, resulting in limited improvements in accuracy across most datasets. These findings support our hypothesis that structured knowledge representations facilitate the precise localization of query-relevant information, enabling more targeted and effective information extraction compared to unstructured contexts.

\section{Related Work}

This section reviews existing research on knowledge conflicts in RAG systems, categorizing the literature into two main areas: impact analysis and resolution strategies.

\subsection{Impact Analysis of Knowledge Conflicts}
Recent studies have extensively explored the influence of knowledge conflicts on the performance of RAG systems~\cite{longpre2021entity,chen2022rich,xie2023adaptive,tan2024blinded,ming2025faitheval}, which primarily highlight differential preferences between the parametric knowledge and retrieved external information. Longpre et al. \cite{longpre2021entity} first expose entity-based knowledge conflicts in question answering, revealing that LLMs tend to rely on parametric memory when retrieved passages are perturbed or contain contradictory information. Chen et al. \cite{chen2022rich} demonstrate that while retrieval-based LLMs predominantly depend on non-parametric evidence when recall is high, their confidence scores fail to reflect inconsistencies among retrieved documents. Xie et al. \cite{xie2023adaptive} find that LLMs are receptive to single external evidence, yet exhibit strong confirmation bias when presented with both supporting and conflicting information. Tan et al. \cite{tan2024blinded} reveal a systematic bias toward self-generated contexts over retrieved ones, attributing this to the higher query-context similarity and semantic incompleteness of retrieved snippets.

Our work aligns with the non-parametric knowledge preference paradigm, aiming to guide LLMs to follow updated and comprehensive external knowledge while correcting for temporal and factual errors within internal memory, thereby generating accurate and trustworthy outputs.

\subsection{Solutions to Knowledge Conflicts}
Current approaches for knowledge conflict resolution can be categorized into token-level and semantic-level methods~\cite{jin2024tug,wang2024astute,bi2025parameters,zhang2025faithfulrag,wang2025accommodate}. Token-level approaches focus on fine-grained intervention during generation. $CD^2$ \cite{jin2024tug} employs attention weight manipulation to suppress parametric knowledge when conflicts are detected. ASTUTE RAG \cite{wang2024astute} utilizes gradient-based attribution to identify and mask conflicting tokens during inference. These methods achieve precise control, but often suffer from computational overhead and lack semantic awareness among generated contents. Semantic-level approaches operate at higher abstraction levels. CK-PLUG \cite{bi2025parameters} develops parameter-efficient conflict resolution through adapter-based architectures that learn to weight parametric versus non-parametric knowledge dynamically. FaithfulRAG \cite{zhang2025faithfulrag} externalizes LLMs' parametric knowledge and aligns it with retrieved context, thereby achieving higher faithfulness without sacrificing accuracy. However, these methods primarily address surface-level conflicts without capturing the underlying factual relationships that drive knowledge inconsistencies.

Different from these approaches, TruthfulRAG leverages structured triple-based knowledge representations to precisely identify and resolve factual-level knowledge conflicts arising from complex natural language expressions, thereby ensuring the reliability and consistency of reasoning.

\section{Conclusion}
In this paper, we introduce TruthfulRAG, the first framework that leverages knowledge graphs to address factual-level conflicts in RAG systems. By integrating systematic triple extraction, query-aware graph retrieval, and entropy-based filtering mechanisms, TruthfulRAG transforms unstructured retrieved contexts into structured reasoning paths that enhance LLMs' confidence in external knowledge while effectively mitigating factual inconsistencies. Our comprehensive experiments demonstrate that TruthfulRAG consistently outperforms existing SOTA methods. These results establish TruthfulRAG as a robust and generalizable solution for improving the trustworthiness and accuracy of RAG systems, with significant implications for knowledge-intensive applications requiring high reliability and precision.


\section*{Acknowledgements}
This work is supported by Funding for Major Science and Technology Breakthrough Projects in Hunan Province (No. 2025QK2009), the National Natural Science Foundation of China No. 62402060, Beijing Natural Science Foundation, No.4244083.

\bibliography{aaai2026}

\appendix

\begin{algorithm}[ht]
\caption{TruthfulRAG: Knowledge Graph-based Conflict Resolution}
\label{alg:truthfulrag}
\begin{algorithmic}[1]
\REQUIRE Query $q$, Retrieved context $C$
\ENSURE Final response $\text{Response}$
\STATE \textbf{// Phase 1: Graph Construction}
\STATE $\mathcal{S} \leftarrow \text{SemanticSegmentation}(C)$
\STATE $\mathcal{T}_{all} \leftarrow \emptyset$
\FOR{$s_i \in \mathcal{S}$}
\STATE $\mathcal{T}_s \leftarrow \text{ExtractTriples}(\mathcal{M}, s)$
\STATE $\mathcal{T}_{all} \leftarrow \mathcal{T}_{all} \cup \mathcal{T}_{i}$
\ENDFOR
\STATE $\mathcal{G} \leftarrow (\mathcal{E}, \mathcal{R}, \mathcal{T}_{all})$

\STATE \textbf{// Phase 2: Graph Retrieval}
\STATE $\mathcal{K}_q \leftarrow \text{ExtractKeyElements}(q)$
\STATE $\mathcal{E}_{imp} \leftarrow \text{TopK}(\{\text{sim}(e, \mathcal{K}_q) : e \in \mathcal{E}\}, k)$
\STATE $\mathcal{R}_{imp} \leftarrow \text{TopK}(\{\text{sim}(r, \mathcal{K}_q) : r \in \mathcal{R}\}, k)$
\STATE $\mathcal{P}_{init} \leftarrow \emptyset$
\FOR{$e \in \mathcal{E}_{imp}$}
\STATE $\mathcal{P}_{2hop} \leftarrow \text{TwoHopTraversal}(e, \mathcal{G})$
\STATE $\mathcal{P}_{init} \leftarrow \mathcal{P}_{init} \cup \mathcal{P}_{2hop}$
\ENDFOR

\STATE \textbf{// Fact-aware path scoring}
\FOR{$p \in \mathcal{P}_{init}$}
\STATE $\text{Ref}(p) \leftarrow \alpha \cdot \frac{|\{e \in p\} \cap \mathcal{E}_{imp}|}{|\mathcal{E}_{imp}|} + \beta \cdot \frac{|\{r \in p\} \cap \mathcal{R}_{imp}|}{|\mathcal{R}_{imp}|}$
\ENDFOR
\STATE $\mathcal{P}_{super} \leftarrow \text{TopK}(\{\text{Ref}(p) : p \in \mathcal{P}_{init}\}, K)$

\STATE \textbf{// Contextualize Reasoning Paths}
\FOR{$p \in \mathcal{P}_{super}$}
\STATE $\mathcal{C}_{path} \leftarrow \text{ExtractSequence}(p)$ \COMMENT{e.g., $e_1 \xrightarrow{r_1} e_2 \cdots$}
\STATE $\mathcal{C}_{entities} \leftarrow \{(e, \mathcal{A}_{e}) : e \in p \cap \mathcal{E}_{imp}\}$
\STATE $\mathcal{C}_{relations} \leftarrow \{(r, \mathcal{A}_{r}) : r \in p \cap \mathcal{R}_{imp}\}$
\STATE $p \leftarrow \mathcal{C}_{path} \oplus \mathcal{C}_{entities} \oplus \mathcal{C}_{relations}$
\ENDFOR

\STATE \textbf{// Phase 3: Conflict Resolution}
\STATE $H_{param} \leftarrow H(P_{param}(ans|q))$
\STATE $\mathcal{P}_{corrective} \leftarrow \emptyset$
\FOR{$p \in \mathcal{P}_{super}$}
\STATE $H_{aug} \leftarrow H(P_{aug}(ans|q, p))$
\STATE $\Delta H_p \leftarrow H_{aug} - H_{param}$
\IF{$\Delta H_p > \tau$}
\STATE $\mathcal{P}_{corrective} \leftarrow \mathcal{P}_{corrective} \cup \{p\}$
\ENDIF
\ENDFOR
\STATE $\text{Response} \leftarrow \mathcal{M}(q \oplus \mathcal{P}_{corrective})$
\RETURN $\text{Response}$
\end{algorithmic}
\end{algorithm}

\section{Problem Statement}

In this section, we formally define the knowledge conflict problem in RAG systems and establish the theoretical foundation for our approach. Let $\mathcal{M}$ denote a LLM equipped with parametric knowledge $\mathcal{K}_p$ acquired during pre-training. Given a query $q$, a standard RAG system retrieves relevant documents $\mathcal{D} = \{d_1, d_2, ..., d_n\}$ from an external knowledge base $\mathcal{K}_e$ and generates a response $y$ by conditioning on both the query and retrieved context.

The knowledge conflict problem arises when there exists a factual inconsistency between the LLMs' parametric knowledge $\mathcal{K}_p$ and retrieved external knowledge $\mathcal{K}_e$ for a given query $q$. Formally, we define a knowledge conflict as follows: 

A knowledge conflict occurs when there exist two factual statements $f_p \in \mathcal{K}_p$ and $f_e \in \mathcal{K}_e$ such that $f_p \not\equiv f_e$, and both statements are relevant to query $q$, where $\not\equiv$ denotes factual inconsistency.

Our objective is to develop a framework that can systematically identify and resolve such knowledge conflicts while maintaining generation quality and ensuring transparent reasoning processes. This entails addressing three key technical challenges: (1) how to effectively represent factual knowledge to facilitate conflict detection; (2) how to retrieve and prioritize relevant factual information for a given query; and (3) how to enable LLMs to make reliable decisions when confronted with conflicting evidence.

\section{Case Study}
To comprehensively demonstrate the efficacy of each component within the TruthfulRAG framework, we conduct a fine-grained case study using a representative instance from the MuSiQue dataset with GPT-4o-mini as the backbone model. The intermediate outputs at each processing stage are detailed in Table~\ref{tab:appendix-example}, which illustrates how TruthfulRAG systematically identifies and resolves knowledge conflicts to achieve consistent and faithful reasoning.
\paragraph{Step 1: Graph Construction} The framework begins by extracting structured knowledge triples from the retrieved context, which contains information like Nuevo Laredo's geographic and administrative characteristics. Through systematic entity-relation-attribute extraction, TruthfulRAG constructs a comprehensive knowledge graph encompassing entities such as "Ciudad Deportiva", "Municipality of Nuevo Laredo", "Nuevo Laredo", and "Sinaloa", along with their intricate relational connections. This structured representation transforms the unstructured natural language text into a semantically enriched knowledge base that facilitates precise factual reasoning.
\paragraph{Step 2: Graph Retrieval} The query-aware graph retrieval algorithm identifies several critical reasoning paths that are semantically aligned with the key information embedded in the user query. For example, these paths systematically trace the ownership hierarchy from Ciudad Deportiva through various intermediate entities, with the most relevant path establishing the connection: "Municipality of Nuevo Laredo" → "Nuevo Laredo" → "Sinaloa". Each reasoning path is enriched with detailed contextual information, including entity attributes and relational descriptions, thereby ensuring semantic coherence and factual completeness.
\paragraph{Step 3: Conflict Resolution} The entropy-based conflict detection mechanism analyzes the model's confidence variations across all retrieved reasoning paths. Notably, the path connecting "Municipality of Nuevo Laredo" to "Sinaloa" exhibits a significant entropy increase, indicating potential factual conflicts with the model's internal parametric knowledge. Through systematic entropy filtering, TruthfulRAG successfully isolates the corrective knowledge path, enabling the model to generate the accurate response "Sinaloa" and effectively resolving the geographical inconsistency present in the original retrieved content.

\section{Algorithm Overview}
Algorithm~\ref{alg:truthfulrag} presents the complete TruthfulRAG framework, which systematically transforms raw retrieval context into structured reasoning paths and improves the factual consistency of model generation through entropy-based confidence filtering.

\section{Additional Experiments}
This section reports four additional experiments, each focusing on a distinct perspective: (1) hyperparameter robustness, (2) significance testing, (3) evaluation on SOTA models, and (4) computational cost analysis. All experiments follow the same implementation settings described in the main paper unless otherwise specified.

\subsection{Hyperparameter Robustness}

To further examine the sensitivity of TruthfulRAG to the entropy threshold $\tau$, we conduct a robustness study by fixing $\tau=1$ across all models, instead of using model-specific thresholds as in the main experiments. This experiment tests whether the conclusions remain stable under a unified hyperparameter configuration.

\paragraph{Setup.}
Following~\citep{bi2025parameters}, the original configuration employs model-specific thresholds ($\tau=1$ for GPT-4o-mini and Mistral-7B-Instruct, $\tau=3$ for Qwen2.5-7B-Instruct) to accommodate the varying conflict sensitivities of different LLMs. In this supplementary experiment, we fix $\tau=1$ for all backbones and re-evaluate TruthfulRAG on four representative benchmarks.

\paragraph{Results and Analysis.}
Table~\ref{tab:tau_unified} presents the results for Qwen2.5-7B-Instruct, comparing the unified-threshold configuration with the original setting. TruthfulRAG achieves comparable performance across all datasets, demonstrating that TruthfulRAG is robust to threshold variations and does not rely on fine-grained hyperparameter tuning, confirming the stability of the method.

\subsection{Significance Testing}

To statistically verify the performance gains of TruthfulRAG over FaithfulRAG, we conduct paired significance testing using GPT-4o-mini as the backbone model. Each dataset is evaluated over 10 independent runs to compute mean, standard deviation, confidence intervals, and $p$-values.

\paragraph{Results and Analysis.}
As shown in Table~\ref{tab:significance}, TruthfulRAG significantly outperforms FaithfulRAG across all datasets, with improvements on four datasets achieving $p<0.05$, confirming that the performance gains are statistically significant rather than attributable to random fluctuations.

\subsection{Evaluation on SOTA LLMs}

To examine the general applicability of TruthfulRAG to stronger LLMs, we evaluate two state-of-the-art LLMs, Gemini-2.5-Flash and Qwen2.5-72B-Instruct, on the RealtimeQA dataset. The results demonstrate that TruthfulRAG continues to yield consistent accuracy improvements even on cutting-edge models.

\paragraph{Results and Analysis.}
TruthfulRAG achieves substantial accuracy improvements on both large-scale LLMs. This result highlights that our method can be effectively extended to LLMs of various architectures and scales.

\subsection{Computational Cost Analysis}

We further analyze the time cost and generated context length of TruthfulRAG compared with baseline RAG systems and FaithfulRAG. All evaluations are performed under identical experimental settings on four datasets.

\paragraph{Results and Analysis.}
As shown in Tables~\ref{tab:timecost} and~\ref{tab:contextlength}, TruthfulRAG introduces moderate computational overhead compared with FaithfulRAG, primarily due to the graph-based reasoning and entropy filtering modules. However, it maintains practical efficiency and compact contextual representations, making it suitable for real-world deployment where both accuracy and trustworthiness are required.

\begin{table*}[ht]
\centering
\renewcommand\arraystretch{1.2}
\setlength{\tabcolsep}{5mm}
\begin{tabular}{c|cccc}
\hline
\textbf{Dataset} & FaithEval & MuSiQue & RealtimeQA & SQuAD \\
\hline
$\tau=3$ (orig.) & 73.2 & 79.1 & 82.3 & 78.7 \\
$\tau=1$ (unified) & 74.2 & 78.7 & 82.4 & 78.8 \\
\hline
\end{tabular}
\caption{Performance of TruthfulRAG under a unified entropy threshold $\tau=1$ using Qwen2.5-7B-Instruct.}
\label{tab:tau_unified}
\end{table*}

\begin{table*}[ht]
\centering
\renewcommand\arraystretch{1.2}
\setlength{\tabcolsep}{3.5mm}
\begin{tabular}{c|ccccc}
\hline
\textbf{Dataset} & FaithfulRAG & TruthfulRAG (mean$\pm$std) & $\Delta$ & 95\% CI & \textit{p} \\
\hline
FaithEval & 67.2 & 69.16$\pm$0.38 & +1.96 & [+1.7,+2.2] & $<$0.001 \\
MuSiQue & 79.3 & 79.71$\pm$0.40 & +0.41 & [+0.1,+0.7] & 0.013 \\
RealtimeQA & 78.8 & 85.00$\pm$0.93 & +6.20 & [+5.5,+6.9] & $<$0.001 \\
SQuAD & 80.8 & 81.30$\pm$0.23 & +0.50 & [+0.3,+0.7] & $<$0.001 \\
\hline
\end{tabular}
\caption{Statistical significance test results based on 10 independent runs with GPT-4o-mini.}
\label{tab:significance}
\end{table*}

\begin{table*}[ht]
\centering
\renewcommand\arraystretch{1.2}
\setlength{\tabcolsep}{6mm}
\begin{tabular}{c|cc}
\hline
\textbf{Method} & \textbf{LLM} & \textbf{RealtimeQA} \\
\hline
\multirow{2}{*}{FaithfulRAG} 
& Gemini-2.5-Flash & 85.84 \\
& Qwen2.5-72B-Instruct & 5.31 \\
\hline
\multirow{2}{*}{TruthfulRAG} 
& Gemini-2.5-Flash & 88.50 \\
& Qwen2.5-72B-Instruct & 84.07 \\
\hline
\end{tabular}
\caption{Performance comparison on RealtimeQA using SOTA LLMs.}
\label{tab:sota}
\end{table*}

\begin{table*}[ht]
\centering
\renewcommand\arraystretch{1.2}
\setlength{\tabcolsep}{6mm}
\begin{tabular}{c|c|cccc}
\hline
\multirow{2}{*}{\textbf{Method}} & \multirow{2}{*}{\textbf{LLM}} & \multicolumn{4}{c}{\textbf{Dataset}} \\
\cline{3-6}
& & FaithEval & MuSiQue & RealtimeQA & SQuAD\\
\hline
\multirow{3}{*}{w/ RAG} 
& Qwen2.5-7B & 0.54 & 0.47 & 0.87 & 0.37 \\
& Mistral-7B & 1.79 & 2.33 & 0.73 & 2.58 \\
& GPT-4o-mini & 0.72 & 0.76 & 0.78 & 0.78 \\
\hline
\multirow{3}{*}{FaithfulRAG} 
& Qwen2.5-7B & 39.79 & 33.91 & 34.19 & 36.75 \\
& Mistral-7B & 54.26 & 44.74 & 47.77 & 49.15 \\
& GPT-4o-mini & 14.56 & 13.18 & 11.51 & 13.91 \\
\hline
\multirow{3}{*}{TruthfulRAG} 
& Qwen2.5-7B & 56.90 & 57.10 & 62.46 & 53.75 \\
& Mistral-7B & 53.58 & 52.42 & 62.12 & 51.30 \\
& GPT-4o-mini & 36.72 & 45.42 & 35.67 & 35.02 \\
\hline
\end{tabular}
\caption{Average time cost (seconds per query).}
\label{tab:timecost}
\end{table*}

\begin{table*}[ht]
\centering
\renewcommand\arraystretch{1.2}
\setlength{\tabcolsep}{6mm}
\begin{tabular}{c|c|cccc}
\hline
\multirow{2}{*}{\textbf{Method}} & \multirow{2}{*}{\textbf{LLM}} & \multicolumn{4}{c}{\textbf{Dataset}} \\
\cline{3-6}
& & FaithEval & MuSiQue & RealtimeQA & SQuAD\\
\hline
\multirow{3}{*}{w/ RAG} 
& Qwen2.5-7B & 374 & 385 & 601 & 259 \\
& Mistral-7B & 374 & 385 & 601 & 259 \\
& GPT-4o-mini & 374 & 385 & 601 & 259 \\
\hline
\multirow{3}{*}{FaithfulRAG} 
& Qwen2.5-7B & 134 & 159 & 155 & 151 \\
& Mistral-7B & 139 & 162 & 158 & 156 \\
& GPT-4o-mini & 136 & 184 & 159 & 169 \\
\hline
\multirow{3}{*}{TruthfulRAG} 
& Qwen2.5-7B & 393 & 287 & 280 & 365 \\
& Mistral-7B & 298 & 149 & 185 & 247 \\
& GPT-4o-mini & 404 & 372 & 255 & 353 \\
\hline
\end{tabular}
\caption{Average generated context length (tokens).}
\label{tab:contextlength}
\end{table*}

\begin{table*}[ht]
\small
\centering
\renewcommand\arraystretch{1.4}
\setlength{\tabcolsep}{2mm}
\begin{tabularx}{\textwidth}{|p{2.2cm}|X|}
    \hline
    \textbf{Query} & What administrative territorial entity is the owner of Ciudad Deportiva located? \\ 
    \hline
    \textbf{Context} & The Municipality of Nuevo Laredo is located in the Mexican state of Sinaloa. Its municipal seat is Nuevo Laredo. The municipality contains more than 60 localities which the most important ones are Nuevo Laredo, El Campanario y Oradel, and Álvarez, the last two being suburbs of the city of Nuevo Laredo…
    \\ 
    \hline
    \textbf{Knowledge Triples} & \textbf{Nodes:}
    \newline "NUEVO LAREDO": "Nuevo Laredo is a city in the Mexican state of Sinaloa, serving as the municipal seat and containing the majority of the municipality's population."...
    \newline "CIUDAD DEPORTIVA": "Ciudad Deportiva, or 'Sports City', is a sports complex in Nuevo Laredo, hosting various sports teams and events."...
    \newline \textbf{Edges:}
    \newline "NUEVO LAREDO" → "SINALOA": "Nuevo Laredo is a city located within the state of Sinaloa, contributing to the state's population and economy."...
    \newline "ESTADIO NUEVO LAREDO" → "TOROS DE NUEVO LAREDO": "Estadio Nuevo Laredo is specifically the baseball park where the Tecolotes de Nuevo Laredo play their home games."...
    \\ 
    \hline
    \textbf{Reasoning Paths} & \textbf{Path 1:} "CIUDAD DEPORTIVA" → "TOROS DE NUEVO LAREDO" → "NUEVO LAREDO MULTIDISCIPLINARY GYMNASIUM" 
    \newline \textbf{Nodes:} 
    \newline \textbf{Edges:} "CIUDAD DEPORTIVA" → "TOROS DE NUEVO LAREDO": "Ciudad Deportiva also serves as the home venue for the Toros de Nuevo Laredo basketball team, hosting their games."...
    \newline "NUEVO LAREDO" → "SINALOA": "Nuevo Laredo is a city located within the state of Sinaloa, contributing to the state's population and economy."...
    \newline 
    \newline 
    \textbf{Path 2:} "MUNICIPALITY OF NUEVO LAREDO" → "NUEVO LAREDO" → "SINALOA"     \newline \textbf{Nodes:} "NUEVO LAREDO": "Nuevo Laredo is a city in the Mexican state of Sinaloa, serving as the municipal seat and containing the majority of the municipality's population."...
    \newline \textbf{Edges:} "NUEVO LAREDO" → "SINALOA": "Nuevo Laredo is a city located within the state of Sinaloa, contributing to the state's population and economy."...
    \newline 
    \newline 
    \textbf{Path 3:} "ESTADIO NUEVO LAREDO" → "TECOLOTES DE NUEVO LAREDO" → "CIUDAD DEPORTIVA" 
    \newline \textbf{Nodes:} "ESTADIO NUEVO LAREDO": "Estadio Nuevo Laredo is a baseball park within Ciudad Deportiva, with a seating capacity of up to 12,000 fans."...
    \newline \textbf{Edges:} "ESTADIO NUEVO LAREDO" → "TOROS DE NUEVO LAREDO": "Estadio Nuevo Laredo is specifically the baseball park where the Tecolotes de Nuevo Laredo play their home games."...
    \\ 
    \hline
    \textbf{Entropy} \newline \textbf{Filtered Path} & "MUNICIPALITY OF NUEVO LAREDO" → "NUEVO LAREDO" → "SINALOA"     \newline \textbf{Nodes:} "NUEVO LAREDO": "Nuevo Laredo is a city in the Mexican state of Sinaloa, serving as the municipal seat and containing the majority of the municipality's population."...
    \newline \textbf{Edges:} "NUEVO LAREDO" → "SINALOA": "Nuevo Laredo is a city located within the state of Sinaloa, contributing to the state's population and economy."...
    \\ 
    \hline
    \textbf{Output} & 
     The context states that Ciudad Deportiva is located in Nuevo Laredo, which is part of the Municipality of Nuevo Laredo in the Mexican state of Sinaloa. Therefore, the administrative territorial entity that owns Ciudad Deportiva is Sinaloa.", "Answer": "\textbf{Sinaloa}" \\
     \hline
\end{tabularx}
\caption{A complete pipeline of our method on MuSiQue using GPT-4o-mini as the backbone model.}
\label{tab:appendix-example}
\end{table*}

\end{document}